\documentclass[11pt]{article}
\usepackage{natbib}
\usepackage{hyperref}
\usepackage{authblk}

\usepackage[inline]{enumitem}
\usepackage{booktabs}
\usepackage[table,dvipsnames]{xcolor}
\usepackage{amssymb}
\usepackage{pifont}
\usepackage{pythonhighlight}
\usepackage{listings}
\usepackage{csquotes}
\usepackage{siunitx}
\usepackage{microtype}  
\usepackage[flushleft]{threeparttable}  
\usepackage{tablefootnote}  
\usepackage{filecontents}

\newcommand{\mytilde}{\raise.17ex\hbox{$\scriptstyle\mathtt{\sim}$}}
\newcommand{\cmark}{\color{ForestGreen}{\ding{51}}}%
\newcommand{\xmark}{\color{BrickRed}{\ding{55}}}%

\definecolor{codegreen}{rgb}{0,0.6,0}
\definecolor{codegray}{rgb}{0.5,0.5,0.5}
\definecolor{codepurple}{rgb}{0.58,0,0.82}
\definecolor{backcolour}{rgb}{0.95,0.95,0.92}
\definecolor{celadon}{rgb}{0.67, 0.88, 0.69}
\definecolor{grannysmithapple}{rgb}{0.66, 0.89, 0.63}
\definecolor{inchworm}{rgb}{0.7, 0.93, 0.36}
\definecolor{lightgreen}{rgb}{0.56, 0.93, 0.56}
\definecolor{palegreen}{rgb}{0.6, 0.98, 0.6}
\definecolor{mossgreen}{rgb}{0.68, 0.87, 0.68}
\definecolor{pastelgreen}{rgb}{0.47, 0.87, 0.47}
\definecolor{pastelred}{rgb}{1.0, 0.41, 0.38}
\definecolor{githubdiffgreen}{rgb}{0.867, 1.0, 0.867}
\definecolor{githubdiffred}{rgb}{1.0, 0.867, 0.867}

\newcommand{\reducedstrut}{\vrule width 0pt height .9\ht\strutbox depth .9\dp\strutbox\relax}
\newcommand{\diffgreen}[1]{%
  \begingroup
  \setlength{\fboxsep}{0pt}%
  \colorbox{githubdiffgreen}{\reducedstrut#1\/}%
  \endgroup
}
\newcommand{\diffred}[1]{%
  \begingroup
  \setlength{\fboxsep}{0pt}%
  \colorbox{githubdiffred}{\reducedstrut#1\/}%
  \endgroup
}

\lstdefinestyle{mystyle}{
    commentstyle=\color{codegreen},
    keywordstyle=\color{magenta},
    numberstyle=\tiny\color{codegray},
    stringstyle=\color{codepurple},
    basicstyle=\ttfamily\footnotesize,
    breakatwhitespace=false,         
    escapeinside={\%*}{*)},            
    breaklines=true,
    captionpos=b,                    
    keepspaces=true,                 
    numbersep=5pt,                  
    showspaces=false,                
    showstringspaces=false,
    showtabs=false,                  
    tabsize=4
}

\title{\texttt{scikit-hubness}: Hubness Reduction and Approximate Neighbor Search}
\author[1]{Roman Feldbauer}
\author[1]{Thomas Rattei}
\author[2]{Arthur Flexer}
\affil[1]{Division of Computational Systems Biology, Department of Microbiology and Ecosystem Science\\
        University of Vienna, Althanstra{\ss}e 14, 1090 Vienna, Austria}
\affil[2]{Austrian Research Institute for Artificial Intelligence (OFAI)\\
        Freyung 6/6/7, 1010 Vienna, Austria}
\affil[ ]{\textit {\{roman.feldbauer,thomas.rattei\}@univie.ac.at},
          \textit {arthur.flexer@ofai.at}}
\date{}                     
\begin{document}
\maketitle

\bibpunct{(}{)}{;}{a}{,}{,}



\begin{abstract}
This paper introduces \texttt{scikit-hubness}, a Python package for efficient
nearest neighbor search in high-dimensional spaces.
Hubness is an aspect of the \textit{curse of dimensionality},
and is known to impair various learning tasks,
including classification, clustering, and visualization.
\texttt{scikit-hubness} provides algorithms for
hubness analysis (``Is my data affected by hubness?''),
hubness reduction (``How can we improve neighbor retrieval in high dimensions?''),
and approximate neighbor search (``Does it work for large data sets?'').
It is integrated into the \texttt{scikit-learn} environment,
enabling rapid adoption by Python-based machine learning
researchers and practitioners.
Users will find all functionality of the \texttt{scikit-learn neighbors} package,
plus additional support for transparent hubness reduction
and approximate nearest neighbor search.
\texttt{scikit-hubness} is developed using several quality assessment tools and principles,
such as PEP8 compliance, unit tests with high code coverage, continuous integration
on all major platforms (Linux, MacOS, Windows), and additional checks by LGTM.
The source code is available at \url{https://github.com/VarIr/scikit-hubness}
under the BSD~3-clause license.
Install from the Python package index
with \texttt{\$ pip install scikit-hubness}.
\end{abstract}

\paragraph{Key words}  hubness, nearest neighbors, curse of dimensionality, scikit-learn, Python

\section{Introduction}
Hubness is a phenomenon in nearest neighbor graphs~\citep{Radovanovic2010},
that is often detrimental to data mining and learning tasks building upon them.
It is an aspect of the \textit{curse of dimensionality},
which describes challenges arising when data are embedded in high dimensions.
Specifically, hubness describes the increasing occurrence of \textit{hubs}
and \textit{antihubs} with increasing data dimensionality:
Hubs are objects, that appear uncommonly often among the nearest neighbors
of others objects, while antihubs are never retrieved as neighbors.
As a consequence, hubs may propagate their information (for example, class labels)
too widely within the neighbor graph, while information from antihubs is depleted.
These semantically distorted graphs can reduce learning performance
in various tasks, such as
classification~\citep{Radovanovic2010},
clustering~\citep{Schnitzer2015},
or visualization~\citep{Flexer2015a}.
Hubness is known to affect a variety of applied learning systems,%
causing---for instance---odd music recommendations~\citep{Schnitzer2012},
or improper transport mode detection~\citep{Feldbauer2018}.

Multiple hubness reduction algorithms have been developed to mitigate these
effects \citep{Schnitzer2012,Flexer2013,Hara2015,Hara2016}.
In a recent survey, we compared these algorithms exhaustively~\citep{Feldbauer2019},
and developed approximate hubness reduction methods with linear time and memory complexity
for the most successful algorithms~\citep{Feldbauer2018}.

In this paper we describe \texttt{scikit-hubness}, which
provides readily available, easy-to-use hubness analysis and reduction
for both machine learning researchers and practitioners.

\section{Software Development and Architecture}
\texttt{scikit-hubness} is a Python package built upon the SciPy stack~\citep{Virtanen2019}
and scikit-learn~\citep{Pedregosa2011}
with cross-platform support for Linux, MacOS, and Windows.
It consists of three sub-packages, that are detailed in the sections below.
Code style and API design is based on the guidelines of \texttt{scikit-learn},
with PEP~8 compliance, NumPy documentation format, and additional criteria,
such as those tested by LGTM.\footnote{Automated code analysis is performed by \url{https://lgtm.com/projects/g/VarIr/scikit-hubness/}.}
Code development is aided by continuous integration on all three platforms,
with an extensive set of tests covering nearly the complete code base.
Source code, issue tracker, quickstart, and additional links are available
at GitHub.\footnote{Source code, issue tracker, and additional links etc.\ can be found at
\url{https://github.com/VarIr/scikit-hubness/}.}
The online documentation is available at Read the Docs.\footnote{Online documentation is available at
\url{https://scikit-hubness.readthedocs.io/}.}

\subsection{\texttt{analysis}: Hubness Analysis}
The \texttt{skhubness.analysis} sub-package allows to determine,
whether data is affected by hubness.
This is often the first step in a user's workflow,
allowing for quick assessment of potential improvements due to hubness reduction.
The package provides several hubness measures,
including long-proven ones, such as the $k$-occurrence skewness~\citep{Radovanovic2010},
as well as recently proposed measures, such as the Robin-Hood index~\citep{Feldbauer2018}.

\subsection{\texttt{reduction}: Hubness Reduction Algorithms}
The \texttt{skhubness.reduction} sub-package provides hubness reduction algorithms.
At the time of writing, the most successful methods~\citep{Feldbauer2019} are available:
mutual proximity, local scaling~\citep{Schnitzer2012},
and DisSim\textsuperscript{Local}~\citep{Hara2016}.
We implement both the exact methods (quadratic complexity),
and their approximations~\citep[linear complexity,][]{Feldbauer2018}.
Neighbor graphs can be hubness-reduced directly with classes from this package,
which may be especially useful for hubness-related research.

\subsection{\texttt{neighbors}: Neighbors-Based Learning}
Practitioners find convenient interfaces to hubness-reduced neighbors-based learning
in the \texttt{skhubness.neighbors} sub-package.
It features all functionality from \texttt{sklearn.neighbors}
and adds support for hubness reduction, where applicable.
This includes, for example,
the supervised \texttt{KNeighborsClassifier} and \texttt{RadiusNeighborsRegressor},
\texttt{NearestNeighbors} for unsupervised learning,
and the general \texttt{kneighbors\_graph}.

\begin{figure*}[t]
    \centering
    \begin{minipage}[t]{.49\textwidth}
    \raggedright
\begin{lstlisting}[language=Python]
from skhubness.data import\
    load_dexter
X, y = load_dexter()

from %*\diffred{sklearn}*).neighbors import\
    KNeighborsClassifier
knn_vanilla = KNeighborsClassifier(
    n_neighbors=5, metric="cosine",
)

from sklearn.model_selection import\
    cross_val_score
acc_vanilla = cross_val_score(
    knn_vanilla, X, y, cv=5)

# Accuracy (vanilla kNN):
print(f"{acc_vanilla.mean():.3f}")
# 0.793
\end{lstlisting}
    \end{minipage}%
    \vline\hfill%
    \noindent
    \begin{minipage}[t]{.49\textwidth}
    \raggedright
\begin{lstlisting}[language=Python]
from skhubness.data import\
    load_dexter
X, y = load_dexter()

from %*\diffgreen{skhubness}*).neighbors import\
    KNeighborsClassifier
knn_mp = KNeighborsClassifier(
    n_neighbors=5, metric="cosine",
    %*
        \diffgreen{hubness="mutual\char`_proximity"}
    *))

from sklearn.model_selection import\
    cross_val_score
acc_mp = cross_val_score(
    knn_mp, X, y, cv=5)

# Accuracy (hubness-reduced kNN)
print(f"{acc_mp.mean():.3f}")
# 0.893
\end{lstlisting}
    \end{minipage}
        \caption{Quickstart example.
                 Left: kNN classification in \texttt{scikit-learn}.
                 Right: Hubness reduction is enabled  in \texttt{scikit-hubness}
                 by providing one additional parameter.%
                 }
        \label{fig:quickstart:example}
\end{figure*}

Time and memory requirements of exact nearest neighbor algorithms
scale quadratically with the number of samples.
In order to support very large data sets,
scikit-hubness provides approximate neighbor search with linear complexity.
Several methods are currently available, including
locality-sensitive hashing and hierarchical navigable small-world graphs
\citep{Aumueller2019,Malkov16}.
Note, that approximate search can be employed in conjunction with
or independently from hubness reduction.
Therefore, \texttt{scikit-hubness} provides powerful neighbor search
in large-scale data, hubness-affected or not.

\section{Installation and Basic Example}
Assuming an existing Python environment, \texttt{scikit-hubness} can be installed
from the Python package index via \texttt{\$ pip install scikit-hubness} at the
command line.

Example~\ref{fig:quickstart:example} shows a quickstart example of text classification.
Adapting \texttt{sklearn}-based nearest neighbor pipelines
to use hubness reduction requires only minor modifications
(see highlighted code).
Given the vast user base of scikit-learn, hubness reduction is now
available to a large number of machine learning researchers
and practitioners with very little transition effort.

\section{Comparison to Similar Software}

\begin{table}[t]
\centering
\footnotesize
\begin{tabular}{@{}llcccccc@{}}
    \toprule
    Tool
        & Platform
            & \begin{tabular}{@{}c@{}}Last \\ update \end{tabular}
                & \begin{tabular}{@{}c@{}}Mutual \\ proximity \end{tabular}
                    & \begin{tabular}{@{}c@{}}Local \\ scaling\end{tabular}
                        & \begin{tabular}{@{}c@{}}DisSim \\ Local\end{tabular}
                            & \begin{tabular}{@{}c@{}}Approximate \\ hubness reduction\end{tabular}
                                & \begin{tabular}{@{}c@{}}scikit-learn \\ compatible\end{tabular}  \\
    \midrule
    HubMiner\tablefootnote{HubMiner is available from \url{http://ailab.ijs.si/tools/hub-miner/}.}
        & Java     & 2015    & \cmark  & \cmark  & \xmark    & \xmark & (N/A) \\

    \rowcolor[gray]{.9}
    Hub-Toolbox\tablefootnote{Hub-Toolbox for MATLAB is available from \url{https://github.com/OFAI/hub-toolbox-matlab}.}
        & Matlab   & 2016    & \cmark  & \cmark  & \xmark    & \xmark & (N/A) \\

    Hub-Toolbox\tablefootnote{Hub-Toolbox for Python is available from \url{https://github.com/OFAI/hub-toolbox-python3}.}
        & Python   & 2019    & \cmark  & \cmark  & \cmark   & \cmark & (partial) \\

    \rowcolor[gray]{.9}
    scikit-hubness
        & Python   & 2019    & \cmark  & \cmark  & \cmark   & \cmark & \cmark \\ 
    \bottomrule
\end{tabular}
\caption{Comparison of hubness reduction software packages}
\label{tab:comparison:software}
\end{table}

Several tools for hubness reduction have previously been developed.
Table~\ref{tab:comparison:software} provides an overview of selected features.
Within the last three years, only the Hub-Toolbox for Python and scikit-hubness
have been actively developed.
Both tools originate from our research groups.
The Hub-Toolbox was released primarily to enable reproducibility for
hubness research performed in our labs.
It features only partial compatibility with scikit-learn,
and is currently lacking a consistent API.
Scikit-hubness succeeds the Hub-Toolbox, and puts a strong focus on usability.
It has been rewritten from scratch in order to provide a more user-friendly,
more easily extensible, fully scikit-learn compatible machine learning package.

\section{Conclusion and Outlook}
Ever since its original release, scikit-learn has become one of the most popular machine learning
frameworks with over \num{65000} dependent repositories on GitHub at the time of writing (August 2019).
Scikit-hubness integrates seamlessly with scikit-learn,
and introduces effective hubness reduction and approximate neighbor search into this ecosystem.
We are confident to provide the most user-friendly hubness analysis and reduction
software package released so far.

Future plans include adaption to significant changes of \texttt{sklearn.neighbors}
likely to be introduced in the next major release:
The \texttt{KNeighborsTransformer} and \texttt{RadiusNeighbors- Transformer}\footnote{%
Upcoming scikit-learn changes: \url{https://github.com/scikit-learn/scikit-learn/pull/10482}}
transform data into sparse neighbor graphs,
which can subsequently be used as input to other estimators.
Hubness reduction and approximate search can then be implemented as \texttt{Transformers}.
This provides the means to turn \texttt{skhubness.neighbors} from a drop-in replacement
of \texttt{sklearn.neighbors} into a scikit-learn plugin,
which will \begin{enumerate*}[label=(\arabic*)]
    \item accelerate development, 
    \item simplify addition of new hubness reduction and approximate search methods, and
    \item facilitate more flexible usage.
\end{enumerate*}

\paragraph{Acknowledgements}We thank Silvan David Peter for testing the software.\\
This research is supported by the Austrian Science Fund (FWF): P27703 and P31988

\vskip 0.2in
\bibliographystyle{plainnat}
\bibliography{scikit-hubness-references}

\end{document}